\pdfoutput=1

\documentclass[runningheads]{llncs}
\usepackage{graphicx}

\usepackage{amsmath,amssymb} % define this before the line numbering.
\usepackage{color}
\usepackage[width=122mm,left=12mm,paperwidth=146mm,height=193mm,top=12mm,paperheight=217mm]{geometry}
\usepackage[labelfont=bf]{caption}
\usepackage{subfigure}
\usepackage{caption}

\usepackage{parskip}
\usepackage{url}
\usepackage{wrapfig}
\usepackage{dblfloatfix}
\usepackage{enumitem}
\usepackage{wrapfig}
\usepackage{microtype}
\usepackage{algorithm}
\usepackage{algorithmic}
\usepackage{booktabs}
\usepackage{multirow}
\usepackage{bm}
\usepackage[flushleft]{threeparttable}

\begin{document}

\newcommand{\pos}{\mathbf{p}}
\newcommand{\posmat}{\mathbf{P}}
\def\eg{\emph{e.g.}}
\def\ie{\emph{i.e.}}
\def\etal{{\em et al.}}

\pagestyle{headings}
\mainmatter

\title{EC-Net: an Edge-aware Point set\\ Consolidation Network} 
% Replace with your title
\makeatletter
\def\blfootnote{\xdef\@thefnmark{}\@footnotetext}
\makeatother

\titlerunning{EC-Net}
% Replace with a meaningful short version of your title

\authorrunning{L. Yu et al.}
% Replace with shorter version of the author list. If there are more authors than fits a line, please use A. Author et al.

%Please write out author names in full in the paper, i.e. full given and family names. 
%If any authors have names that can be parsed into FirstName LastName in multiple ways, please include the correct parsing, in a comment to the volume editors:
%\index{Lastnames, Firstnames}
%(Do not uncomment it, because you may introduce extra index items if you do that, we will use scripts for introducing index entries...)

\author{Lequan Yu$^{1,3\,\star}$ , ~~Xianzhi Li$^{1\,\star}$, ~~Chi-Wing Fu$^{1,3}$, \\Daniel Cohen-Or$^{2}$, ~~Pheng-Ann Heng$^{1,3}$}
%\index{Yu, Lequan}
%\index{Li, Xianzhi}
%\index{Fu, Chi-Wing}
%\index{Cohen-Or, Daniel}
%\index{Heng, Pheng-Ann}

\institute{\textsuperscript{1}The Chinese University of Hong Kong\hspace{4mm}\textsuperscript{2}Tel Aviv University\\
	\textsuperscript{3}Shenzhen Key Laboratory of Virtual Reality and Human Interaction Technology, Shenzhen Institutes of Advanced Technology, Chinese Academy of Sciences, China\\
	\email{ \{lqyu,xzli,cwfu,pheng\}@cse.cuhk.edu.hk \hspace{4mm} dcor@mail.tau.ac.il} }

\blfootnote{$^\star$ indicates equal contributions.}
\maketitle

\begin{abstract}
	
	Point clouds obtained from 3D scans are typically sparse, irregular, and noisy, and required to be consolidated.
	In this paper, we present the first deep learning based {\em edge-aware} technique to facilitate the consolidation of point clouds.
	We design our network to process points grouped in local patches, and train it to learn and help consolidate points, deliberately for edges.
	To achieve this, we formulate a regression component to simultaneously recover 3D point coordinates and point-to-edge distances from upsampled features, and an edge-aware joint loss function to directly minimize distances from output points to 3D meshes and to edges.
	Compared with previous neural network based works, our consolidation is {\em edge-aware\/}.
	During the synthesis, our network can attend to the detected sharp edges and enable more accurate 3D reconstructions.
	Also, we trained our network on virtual scanned point clouds, demonstrated the performance of our method on both synthetic and real point clouds, presented various surface reconstruction results, and showed how our method outperforms the state-of-the-arts.
	
\end{abstract}

\keywords{point cloud, learning, neural network, edge-aware}

%%%%%%%%%%%%%%%%%%%%%%%%%%%%%%%%%%%%%%%%%%%%%%%%%%%%%%%%%%%%%%%%
\section{Introduction}
\label{sec:introduction}

Point cloud consolidation is \textit{a process of ``massaging'' a point set into a surface}~\cite{alexa2003computing}, for enhancing the surface reconstruction quality. 
In the past two decades, a wide range of techniques have been developed to address this problem, including denoising, completion, resampling, and many more.
However, these techniques are mostly based on {\em priors\/}, such as piecewise smoothness.
Priors are typically over-simplified models of the actual geometry behavior, thus the prior-based techniques tend to work well for specific class of models rather than being general.

To implicitly model and characterize the geometry behavior, one common way is to take a data-driven approach and model the complex behavior using explicit examples.
Data-driven surface reconstruction techniques~\cite{gal2007surface,sung2015data,xu2017data,remil2017surface} are based on matching local portions (often denoted as patches) to a set of examples.
Particularly, the emergence of neural networks and their startling performance provide a new means for 3D reconstruction from point sets by data-driven learning~\cite{guerrero2017pcpnet,yu-2018-pu-net,Groueix2018AtlasNet}.
One of the main limitations of these neural network based methods is that they are oblivious to sharp features on 3D objects, where undersampling problems are typically more severe, making it challenging for an accurate object reconstruction.

\begin{figure}[t]
	\centering
	\includegraphics[height=4.5cm]{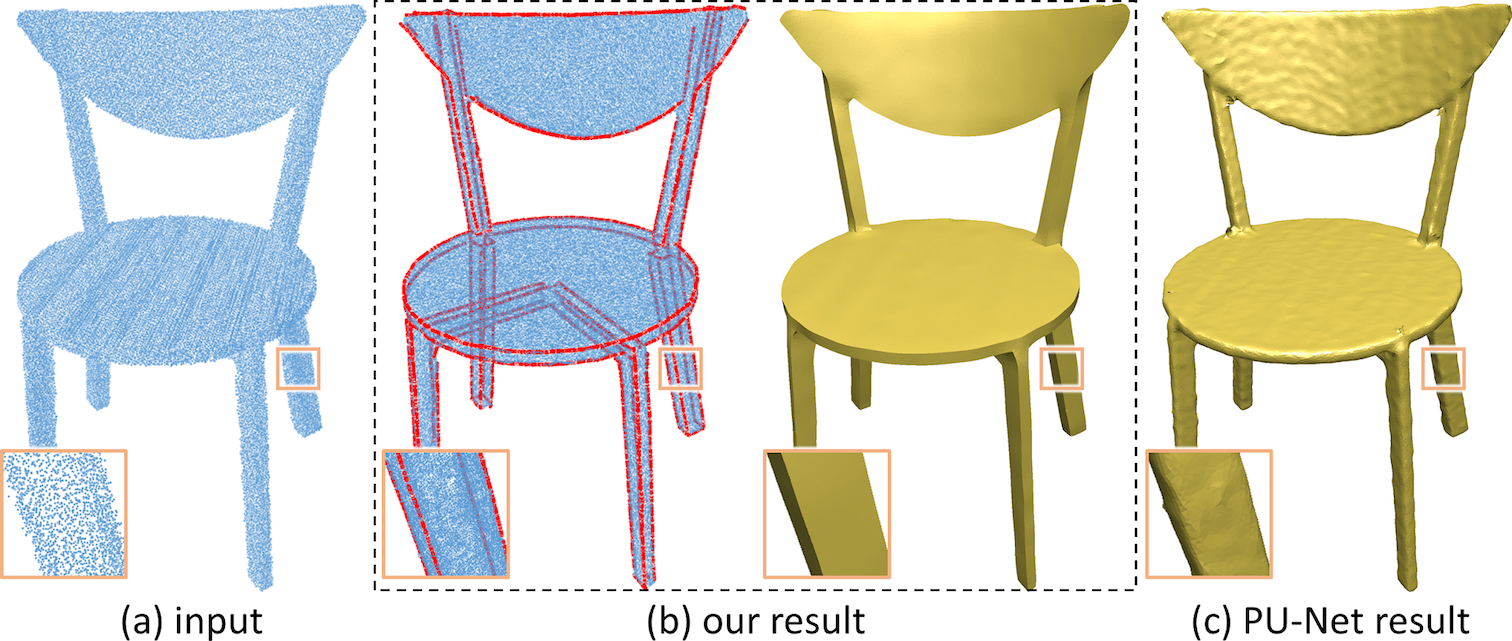}
	\caption{Given a point cloud (a) with noisy samples in inhomogeneous distribution, our method consolidates it and reconstructs a plausible surface (b).
		Compared with PU-Net (c), our method is edge-aware and can preserve sharp features.}
	\label{fig:teaser}
\end{figure}

In this paper, we present the first {\em edge-aware consolidation network\/}, namely EC-Net, for point cloud consolidation.
The network is designed and trained, such that the output points admit to the surface characteristic of the 3D objects in the training set.
More importantly, our method is {\em edge-aware\/}, in the sense that the network learns the geometry of edges from the training set, and during the test time, it identifies edge points and generates more points along the edges (and over the surface) to facilitate a 3D reconstruction that preserves sharp features.

Generally speaking, scanned point sets are irregular and non-uniform, and thus, do not lend themselves to be learned by common convolutional neural networks (CNN).
Inspired by PointNet~\cite{qi2017pointnet}, we directly process 3D points by converting their coordinates into deep features and producing more points by feature expansion~\cite{yu-2018-pu-net}.
Then, for efficient learning of the edges, we design our network to process points grouped as local patches in the point cloud.
To do so, we develop a patch extraction scheme that solely works on points, so that we can extract patches of points for use consistently in both training and testing phases.

In addition, to train the network to be edge-aware, we associate edge and mesh triangle information with the training patches, and train the network to learn features from the patches by regressing point-to-edge distances and then the point coordinates.
More importantly, we design a novel edge-ware joint loss function that can be efficiently computed for directly comparison between the output points and ground truth 3D meshes.
Our loss function encourages the output points to be located close to the underlying surface and to the edges, as well as distributed more evenly on surface.
Then in the inference phase, the network can generate and find output points close to the edges.
Since it is difficult to annotate edges directly in real scanned point clouds, we train our network on synthesized virtual scanned point clouds, and show the performance of our method on both real and virtual scanned point clouds.
By using our trained network, we show through various experiments that we can improve not only the point cloud consolidation results (see Figures~\ref{fig:teaser}(b) \& (c)), but also the surface reconstruction quality, compared to various state-of-the-art methods.
All the code is available at the project webpage\footnote{\url{https://yulequan.github.io/ec-net/index.html}}.

{\bf Related works.} \
Consolidating scanned data and imperfect point clouds has been an active research area since the early 90's~\cite{hoppe1992,turk-levoy-1994,amenta-1998}.
We briefly review some traditional geometric works and then discuss some recent related works that employ neural networks.
For a more comprehensive survey, please refer to~\cite{berger2017survey}.

\textit{Point cloud consolidation.} \
Early works in this area assumed smooth surface~\cite{alexa2003computing,lipman2007parameterization,huang2009consolidation}.
In~\cite{lipman2007parameterization}, the parameterization-free local projection operator (LOP) was devised to enhance the point set quality. 
However, these methods are oblivious to sharp edges and corners.
To consolidate a point set in an edge-aware manner, some methods detected/sensed the sharp edges and arranged points deliberatively along edges to preserve their sharpness~\cite{pauly2003shape,guennebaud2004real,fleishman2005robust,oztireli2009feature}.
Huang~\etal~\cite{huang2013edge} developed the edge-aware resampling (EAR) algorithm; it computes reliable normals away from edges and then progressively upsamples points towards the surface singularities.
Despite its promising results, EAR depends on the accuracy of the given/estimated normal.
Preiner~\etal~\cite{preiner2014continuous} developed CLOP, a continuous version of the LOP, for fast surface construction using the Gaussian mixture model to describe the point cloud density.
To sum up, these geometric approaches either assume strong priors or rely on extra geometric attributes for upsampling point sets.

\textit{Neural networks for mesh and point cloud processing.} \
Motivated by the promising results that deep learning methods have achieved for image and video problems, there has been increasing effort to leverage neural networks for geometry and 3D shape problems.
To do so, early works extracted low-level geometric features as inputs to CNNs~\cite{guo20153d,boulch2016deep}.
Other works converted the input triangular meshes or point clouds to regular voxel grids~\cite{Qi_2016_CVPR,dai2017complete,HRSC,wang2017cnn,Riegler2017OctNet,Liu_2017_ICCV} for CNN to process.
However, pre-extracting low-level features may bring bias, while a volume representation demands a high computational cost and is constrained by its resolution.

Recently, point clouds have drawn more attention, and there are some works to utilize neural networks to directly process point clouds. 
Qi~\etal~\cite{qi2017pointnet} firstly developed the PointNet, a network that takes a set of unordered points in 3D as inputs and learns features for object classification and segmentation.
Later, they proposed the PointNet++ to enhance the network with a hierarchical feature learning technique~\cite{qi2017pointnet++}.
Subsequently, many other networks have been proposed for high-level analysis problems with point clouds~\cite{hua2017pointwise,klokov2017escape,landrieu2017large,xu2017pointfusion,wang2017sgpn,yang2017foldingnet,qi2017frustum,li2018pointcnn,wang2018dynamic,su2018splatnet}.
However, they all focus on analyzing global or mid-level attributes of point clouds.
In some other aspects, Guerrero~\etal~\cite{guerrero2017pcpnet} proposed a network to estimate the local shape properties in point clouds, including normal and curvature.
3D reconstruction from 2D images has also been widely studied~\cite{Groueix2018AtlasNet,Fan_2017_CVPR,lin2017learning}.
Our work is most related to PU-Net~\cite{yu-2018-pu-net}, which presented a network to upsample a point set. However, our method is edge-aware, and we extract local patches and train the network to learn edges in patches with a novel edge-aware joint loss function.

%%%%%%%%%%%%%%%%%%%%%%%%%%%%%%%%%%%%%%%%%%%%%%%%%%%%%%%%%%%%%%%%
\section{Method}
\label{sec:overview}

\begin{figure*}[t]
	\centering
	\includegraphics[width=0.95\linewidth]{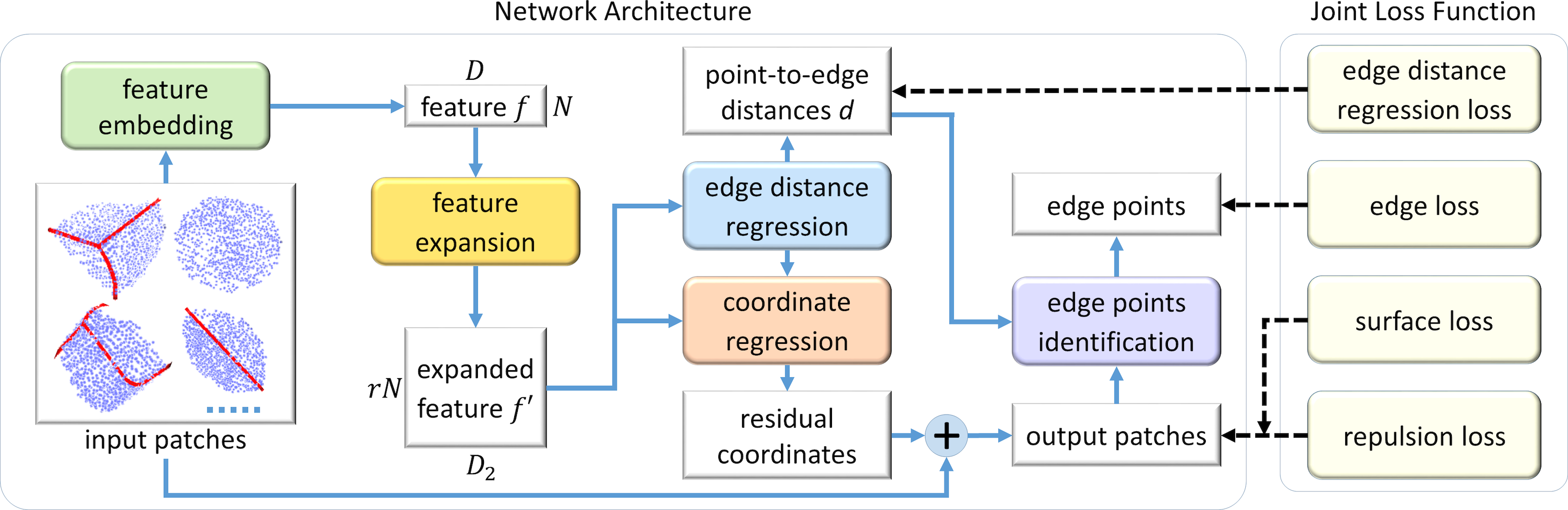}
	\caption{The pipeline of EC-Net.
		For each point in an input patch, we first encode its local geometry into a feature vector $f$ (size: $N \times D$) using PointNet++, and expand $f$ into $f'$ (size: $rN \times D_2$) using a feature expansion mechanism.
		Then, we regress the residual point coordinates and also the point-to-edge distances ($d$) from the expanded features, and form the output point coordinates by adding the original point coordinates to the residual.
		Finally, the network identifies points on edges and yields output points.
		The network was trained with an edge-aware joint loss function that has four terms; see the yellow boxes on the right.}
	\label{fig:overview}
\end{figure*}

In this section, we first present the training data preparation (Sec.~\ref{sec:traindatapreparation}) and the EC-Net architecture (Sec.~\ref{sec:networkarchitecture}).
Then, we present the edge-aware joint loss function (Sec.~\ref{sec:objectivefunction}) and the implementation details (Sec.~\ref{sec:trainingandsynthesis}).
Figure~\ref{fig:overview} shows the pipeline of EC-Net; see the supplemental material for the detailed architecture.

\subsection{Training data preparation}
\label{sec:traindatapreparation}

\begin{figure}[!t]
	\centering
	\includegraphics[width=0.75\linewidth]{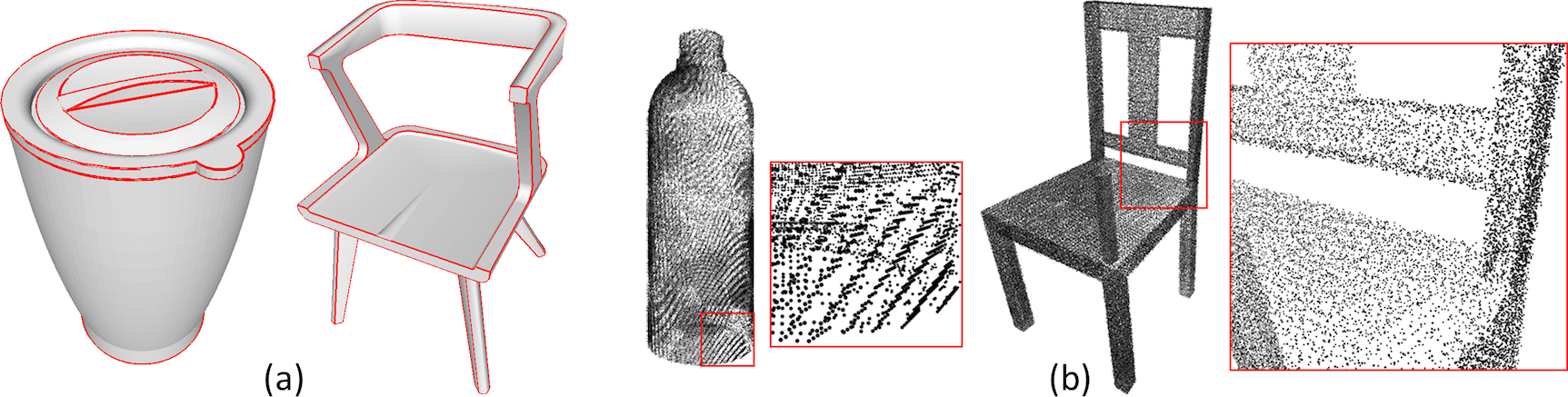}
	\caption{Example annotated edges (in red) on some of our collected 3D meshes (a).
		Example point clouds produced from our virtual 3D scans (b).
		The point density varies and the zoom-in windows also reveal the synthetic noise.}
	\label{fig:annotate_edges}
\end{figure}
We train our network using point clouds synthesized from 3D objects, so that we can have ground truth surface and edge information.
To start, we collect 3D meshes from ShapeNet~\cite{chang2015shapenet} and other online repositories, including simple 3D shapes, mechanical parts, and everyday objects such as chairs.
Since we train the network with patches as inputs, we prepare a large amount of patches on the 3D meshes and do not require many meshes.
Moreover, we manually sketch polylines on each 3D mesh to annotate sharp edges on the meshes; see Figure~\ref{fig:annotate_edges}(a).

{\bf Virtual scanning.} \
To obtain point clouds from the 3D mesh objects, we use the following virtual scanning procedure.
First, we normalize the mesh to fit in $[-1,+1]^3$, and evenly arrange a circle of 30 virtual cameras ($50^\circ$ field of view) horizontally around (and to look at) the object.
We then put the cameras two units from the object center and randomly perturb the camera positions slightly upward, downward or sideway.
After that, we produce a point cloud for each camera by rendering a depth image of the object, adding quantization noise (see Sec.~\ref{sec:experiments}) to the depth values and pixel locations, and backprojecting each foreground pixel to obtain a 3D point sample.
Then, we can compose the 3D point clouds from different cameras to obtain a virtual scanned data.
Such sampling procedure mimics a real scanner with surface regions closer to the virtual camera receiving more point samples; see Figure~\ref{fig:annotate_edges}(b) for two example results.

\begin{figure}[!t]
	\centering
	\includegraphics[width=0.68\linewidth]{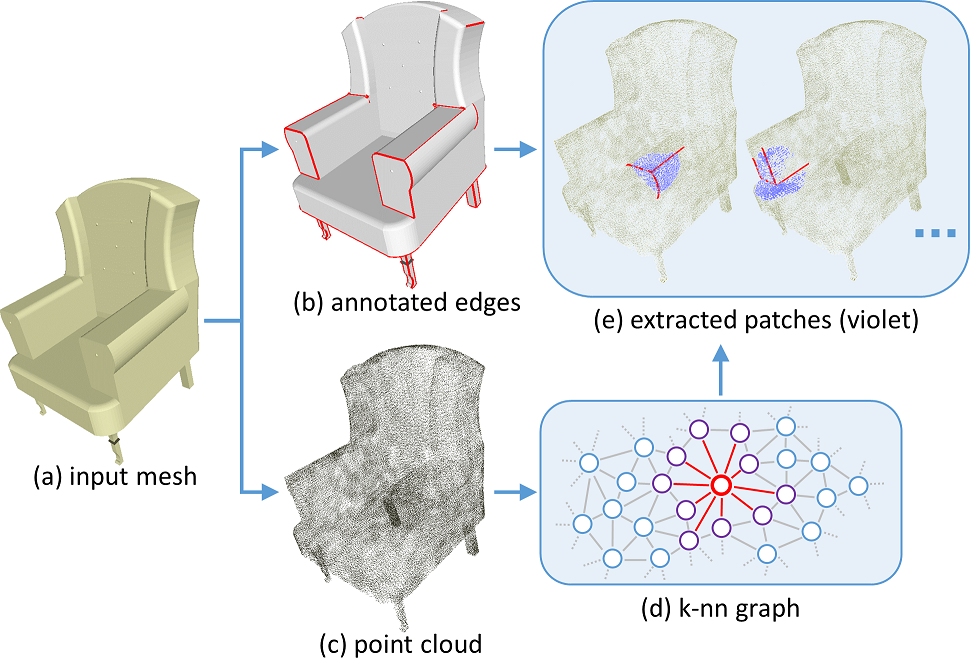}
	\caption{Procedure to extract patches (a local group of points) from a point cloud; note that (a) and (b) are available only in the training (but not inference) phase.}
	\label{fig:prepare_patches}
	\vspace*{-1mm}
\end{figure}

{\bf Patch extraction.} \
From a point cloud (see Figure~\ref{fig:prepare_patches}(c)), we aim to extract local groups of points (patches), such that the points in a patch are {\em geodesically\/} close to one another over the underlying surface.
This is very important, since using Euclidean distances to select points could lead to points on opposite sides of a thin surface, e.g., see the thin plates in the chair shown in Figure~\ref{fig:annotate_edges}(b).
Compared with~\cite{yu-2018-pu-net}, our patch extraction procedure directly operates on point clouds, {\em not\/} meshes, so we need a {\em consistent\/} extraction procedure for {\em both\/} network training and inference, where ground truth meshes are not available during the inference.

To this end, we first construct a weighted graph by considering each point as a node and creating an edge from each point to its k-nearest neighboring (k-nn) points, where k=10; see Figure~\ref{fig:prepare_patches}(d).
The edge weight is set as the Euclidean distance between the two points.
Then, we randomly select $m$$=$$100$ points as the patch centroids; from each selected point, we use the Dijkstra algorithm to find the 2048 nearest points in terms of shortest path distances in the graph.
Hence, we can find points that are approximately within a geodesic radius from the centroid.
Further, we randomly select $\hat{N}$$=$$1024$ points out of the 2048 points to introduce randomness into the point distribution, and normalize the 3D coordinates of the points to have zero mean inside a unit ball.
For patches used for training, we also find the associated mesh triangles and annotated edge segments near the patches as the ground truth information for training the network; see Figure~\ref{fig:prepare_patches}(e).

\subsection{Edge-aware Point set Consolidation Network}
\label{sec:networkarchitecture}
In this subsection, we present the major components of EC-Net; see Figure~\ref{fig:overview}.

{\bf Feature embedding and expansion.} \
This component first maps the neighboring information (raw 3D coordinates of nearby points) around each point into a feature vector using PointNet++~\cite{qi2017pointnet++} to account for the fact that the input points are irregular and unordered.
The output is a $D$-dimensional multi-scale feature vector for each input point, where $D$ is 256 in our experiments.
In this step, we make the following adaptation for our problem.
By design, PointNet++ processes a full point cloud of an object, while EC-Net processes local patches.
Since patches have open boundary, points near the boundary have neighbors mostly on one of its side only, so we found that the extracted features of these points are less accurate.
Hence, out of the $\hat{N}$ feature vectors, we retain the $N$$=$$\frac{\hat{N}}{2}$ feature vectors (denoted as $f$) corresponding to points closer to the patch centroid.
Next, the component synthesizes points by expanding features directly in feature space using the feature expansion module in~\cite{yu-2018-pu-net}, since points and features should be interchangeable.
After this module, feature $f$ (dimension: $N\times D$) are expanded to be $f'$ (dimension: $rN \times D_2$), where $r$ is the upsampling rate and $D_2$ is the new feature dimension, which is set as $128$; see again Figure~\ref{fig:overview}.

{\bf Edge distance regression.} \
This component regresses a point-to-edge distance for each expanded feature (or point, equivalently) later for edge points identification.
The regressed distance is an estimated shortest distance from the output point to the nearest annotated edge segment among all annotated edge segments associated with the patch.
To do this, we extract a distance feature $f_{dist}$ from the expanded feature $f'$ via a fully connected layer, and then regress the point-to-edge distance $d$ from $f_{dist}$ via another fully connected layer.
We do this in two steps, so that we can feed $f_{dist}$ also to the coordinate regression component.

{\bf Coordinate regression.} \
This component reconstructs the 3D coordinates of the output points; see Figure~\ref{fig:overview}.
First, we concatenate the expanded feature $f'$ with the distance feature $f_{dist}$ (from previous component) to form another feature, since $f_{dist}$ contains certain point-to-edge distance information.
Then, we regress the point coordinates from the concatenated feature by applying two fully connected layers. 
Note that we only regress the residual 3D coordinates, and the network output 3D coordinates of the output points by adding the original 3D coordinates of the input points to the regressed residual 3D coordinates. 

{\bf Edge points identification.} \
Denoting $d_i$ as the regressed point-to-edge distance of output point $x_i$, we next find a subset of output points, namely {\em edge points} (denoted as $\mathcal{S}_{\Delta_d}$ with threshold $\Delta_d$) that are near the edges: 
$\mathcal{S}_{\Delta_d}=\{x_i\}_{d_i<\Delta_d}$.
Note that this component is performed in both training and inference phases.

\subsection{Edge-aware joint loss function}
\label{sec:objectivefunction}

The loss function should encourage the output points to be
(i) located close to the underlying object surface,
(ii) edge-aware (located close to the annotated edges), and
(iii) more evenly distributed on the object surface.
To this end, we guide the network's behavior by designing an {\em edge-aware joint loss function\/} with the following four loss terms (see also the rightmost part of Figure~\ref{fig:overview}):

{\bf Surface loss} encourages the output points to be located close to the underlying surface.
When extracting each training patch from the input point clouds, we find triangles and edge segments associated with the patch; see Figure~\ref{fig:prepare_patches}.
Hence, we can define surface loss using the minimum shortest distance from each output point $x_i$ to all the mesh triangles $T$ associated with the patch:
$d_T(x_i,T) = \min_{t \in T} \ {d_t(x_i, t)}$,
where $d_t(x_i,t)$ is the shortest distance from $x_i$ to triangle $t \in T$.
It is worth noting that to compute $d_t$ in 3D, we need to consider seven cases, since the point on $t$ that is the closest to $x_i$ may be located at triangle vertices, along triangle edges, or within the triangle face.
Experimentally, we found that the algorithm~\cite{eberly1999distance} for calculating $d_t$ can be implemented using TensorFlow to automatically calculate the gradients when training the network.
With $d_T$ computed for all the output points, we can sum them up to compute the surface loss:
\begin{equation}
L_{surf} = \frac{1}{\tilde{N}}\sum_{1\leq i \leq \tilde{N}}{d_T^2(x_i,T)} \ ,
\end{equation}
where $\tilde{N}=rN$ is the number of output points in each patch.

{\bf Edge loss} encourages the output points to be edge-aware, i.e., located close to the edges.
Denoting $E$ as the set of annotated edge segments associated with a patch, we define edge loss using the minimum shortest distance from each \emph{edge point} to all the edge segments in the patch:
$d_E(x_i,E) = \min_{e \in E} \ {d_e(x_i,e)} $,
where $d_e(x_i,e)$ is the shortest distance from edge point $x_i$ to any point on edge segment $e \in E$.
Again, we implement the algorithm in~\cite{eberly1999distance2} to calculate $d_e$ for different shortest distance cases using TensorFlow to automatically calculate the gradients.
Then, we sum up $d_E$ for all the edge points and obtain the edge loss:
\begin{equation}
L_{edge}
\ = \
\frac{\sum_{x_i\in \mathcal{S}_{\Delta_d}}{d_E^2(x_i,E)}}{|\mathcal{S}_{\Delta_d}|} \ , \ \ \text{where} \ \mathcal{S}_{\Delta_d} \ \text{is the edge point set} \ .
\end{equation}

{\bf Repulsion loss} encourages the output points to be more evenly distributed over the underlying surface.
Given a set of output points $x_i, i=1 ... \tilde{N}$, it is defined as
\begin{equation}
L_{repl}
\ = \
\frac{1}{\tilde{N}\cdot K} \sum_{1 \leq i \leq \tilde{N}} \ \sum_{i' \in \mathcal{K}(i)}{\eta( \ || \ x_{i'} \ - \ x_i \ || \ )} \ ,
\end{equation}
where
$\mathcal{K}(i)$ is the set of indices for the $K$-nearest neighborhood of $x_i$ (we set $K$=4),
$||\cdot||$ is the L2-norm,
and
$\eta(r) = max(0, h^2-r^2)$ is a function to penalize $x_i$ if it is too close to some other nearby points, where $h$ is empirically set as 0.03 (which is the mean separation distance between points estimated from the number of points and bounding box diagonal length according to~\cite{huang2013edge}).
It is worth noting that we only want to penalize $x_i$ when it is too close to some neighborhood points, so we only consider a few nearest neighboring points around $x_i$; moreover, we remove the repulsion effect when the point-to-point distance is above $h$.

{\bf Edge distance regression loss} aims to guide the network to regress the point-to-edge distances $d$ for the $rN$ output points; see Figure~\ref{fig:overview}.
Considering that it is difficult for the network to regress $d$, since not all output points are actually close to the annotated edges.
Hence, we design a truncated regression loss:
\begin{equation}
L_{regr}
\ = \
\frac{1}{\tilde{N}} \sum_{1 \leq i \leq \tilde{N}} \ {\big[ \ \mathcal{T}_b( d_E(x_i,E) ) \ - \ \mathcal{T}_b( d_i ) \ \big]^2} \ ,
\end{equation}
where
$\mathcal{T}_b(x) = max (0, min(x,b))$ is a piecewise linear function with parameter $b$.
Empirically, we found the network training not sensitive to $b$, and set it as 0.5.

{\bf End-to-end training.} \
When training the network, we minimize the combined edge-aware joint loss function below with balancing weights $\alpha$ and $\beta$:
\begin{equation}
\label{eq:joint-loss}
\mathcal{L} \ = \
L_{surf}
\ + \
L_{repl}
\ + \
\alpha L_{edge}
\ + \
\beta L_{regr} \ .
\end{equation}
In our implementation, we set $\alpha$ and $\beta$ as $0.1$ and $0.01$, respectively.

\subsection{Implementation Details}
\label{sec:trainingandsynthesis}

{\bf Network training.} \
Before the training, each input patch is normalized to fit in $[-1,1]^3$.
Then, we augment each patch on-the-fly in the network via a series of operators:
a random rotation,
a random translation in all dimensions by -0.2 to 0.2,
a random scaling by 0.8 to 1.2,
adding Gaussian noise to the patch points with $\sigma$ set as $0.5\%$ of the patch bounding box size, and
randomly shuffling the ordering of points in the patch.
We implement our method using TensorFlow, and train the network for 200 epochs using the Adam~\cite{kingma2014adam} optimizer with a minibatch size of 12 and a learning rate of 0.001.
In our experiments, the default upsampling rate $r$ is set as 4.
For threshold $\Delta_d$, we empirically set it as $0.15$, since it is not sensitive to slight variation in our experiments.
Overall, it took around 5 hours to train the network on an NVidia TITAN Xp GPU.

{\bf Network inference.} \
We apply a trained network to process point clouds also in a patch-wise manner.
To do so, we first find a subset of points in a test point cloud, and take them as centroids to extract patches of points using the procedure in Sec.~\ref{sec:traindatapreparation}.
For the patches to distribute more evenly over the point cloud (say with $N_{pt}$ points), we use farthest point sampling to randomly find $M=\beta \frac{N_{pt}}{N}$ points in the test point cloud with parameter $\beta$, which is empirically set as three.
Hence, each point in the point cloud should appear roughly in $\beta$ different patches on average.
After extracting the patches, we feed them into the network and apply the network to produce 3D coordinates and point-to-edge distances, as well as to identify edge points (see Sec.~\ref{sec:objectivefunction}).
Unlike the training phase, we set a smaller $\Delta_d$, which is $0.05$.
We use a larger $\Delta_d$ in the training because training is an optimization process, where we want the network to consider more points to learn to identify the points near the edges.

{\bf Surface reconstruction.} \
First, we build a k-nn graph for the output points from network.
Then, we filter edge points by fitting line segments using RANSAC, and filter surface points (not near edges points) by finding small groups of nearby points in the k-nn graph in an edge-stopping manner and fitting planes using PCA.
Edge stopping means we stop the breath-first growth at edge points; this avoids including irrelevant points beyond the edges.
These steps are iterated several times.
Lastly, we fill the tiny gap between edge and surface points by including some original points in the gap, and by applying dart throwing to add new points.
To further reconstruct the surface, we follow the procedure in EAR~\cite{huang2013edge} to downsample the point set and compute normals, use ball pivoting~\cite{bernardini1999ball} or screened Poisson surface reconstruction~\cite{kazhdan2013screened} to reconstruct the surface, and use a bilateral normal filtering~\cite{zheng2011bilateral} to clean the resulting mesh.

%%%%%%%%%%%%%%%%%%%%%%%%%%%%%%%%%%%%%%%%%%%%%%%%%%%%%%%%%%
\begin{figure*}[!t]
	\centering
	\includegraphics[width=0.98\linewidth]{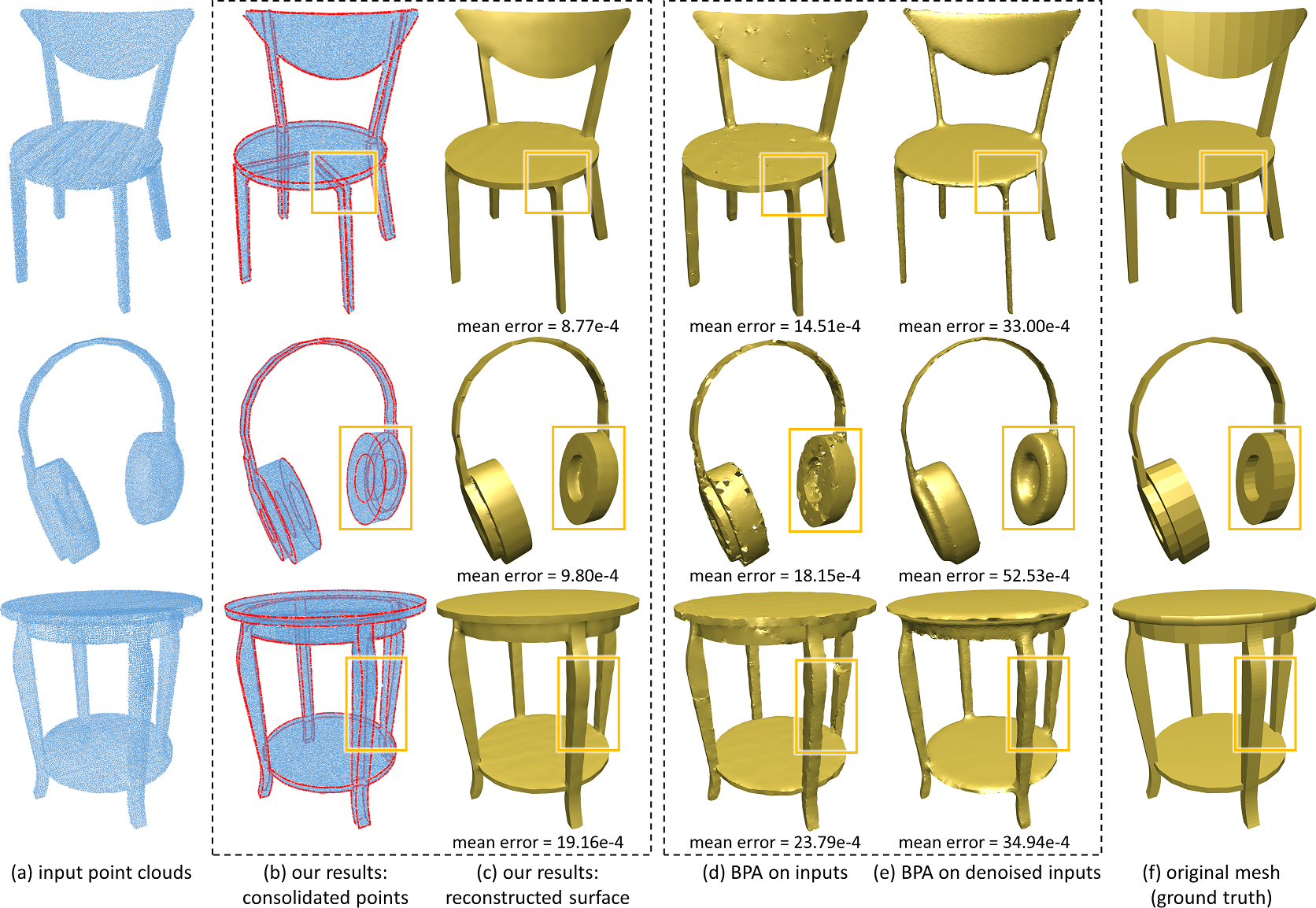}
	\caption{Our method produces consolidated point clouds (b) that lead to higher quality surface reconstruction results (c).
		Predicted edge points are shown in red.}
	\label{fig:surfacere_construction}
\end{figure*}

\section{Experiments}
\label{sec:experiments}

\begin{figure*}[t]
	\centering
	\includegraphics[width=0.95\linewidth]{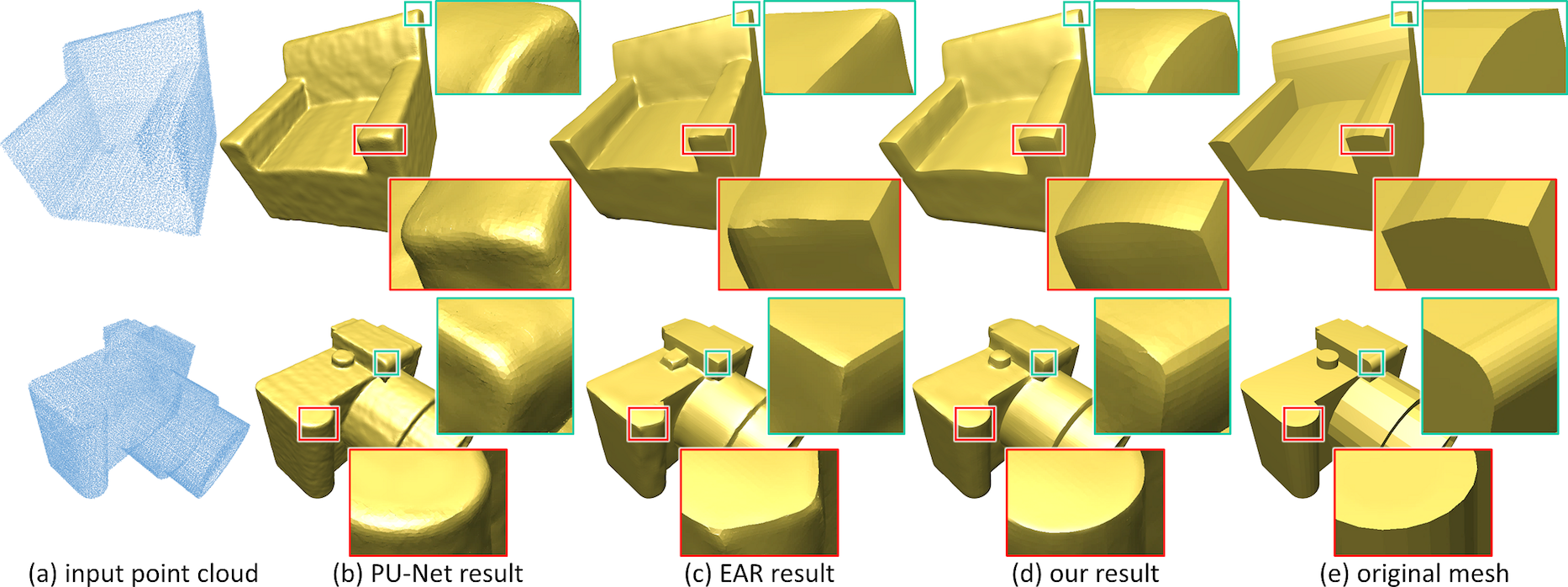}
	\caption{Comparing surface reconstruction results produced by using PU-Net (b), EAR (c), and our method (d) to consolidate points with the original meshes (e).}
	\label{fig:comparison1}
\end{figure*}

\begin{figure*}[t]
	\centering
	\includegraphics[width=0.95\linewidth]{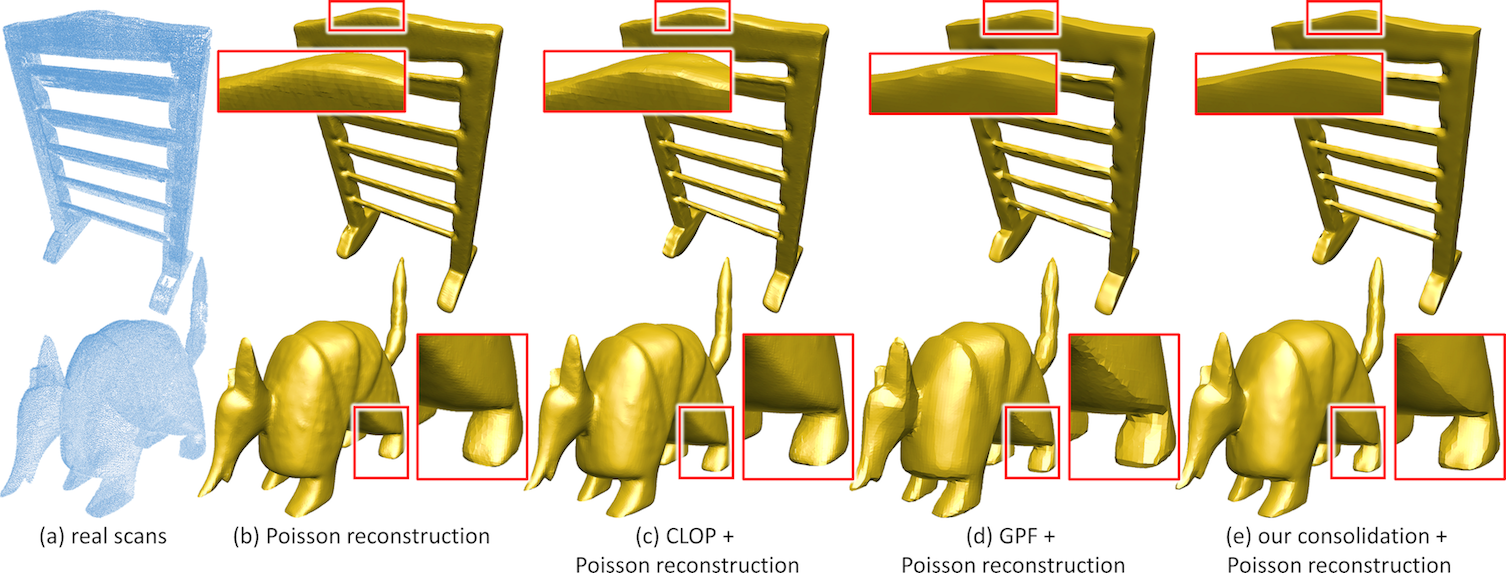}
	\caption{Comparing reconstruction results on real scans produced by direct reconstruction (b) and with consolidation: CLOP, GPF, and our method (c-e).}
	\label{fig:comparison2}
\end{figure*}

%%%%%%%%%%%%%%%%%%%%%%%%%%%%%%%%%%%%%%%%%%%%%%
{\bf Dataset overview.} \
\label{sec:dataset}
Since most models in~\cite{yu-2018-pu-net} are manifolds without sharp edges, 
we collected 24 CAD models and 12 everyday objects as our training data set, and manually annotate sharp edges on them; see supplemental material.
Then, we randomly crop 2,400 patches from the models (see Figure~\ref{fig:overview}) to train our network; see the procedure in Sec.~\ref{sec:traindatapreparation}.
To perform the experiments presented in this section, we do not reuse the models in the training data set but download additional 3D models from ShapeNet~\cite{chang2015shapenet}.
For each testing model, we also use the procedure in Sec.~\ref{sec:traindatapreparation} to generate the virtual scanned point clouds as input.

{\bf Surface reconstruction results.} \
We demonstrate the quality of our method by applying it to consolidate point sets and reconstruct surfaces.
Figure~\ref{fig:surfacere_construction}(a) shows the input point clouds generated from the testing models shown in Figure~\ref{fig:surfacere_construction}(f), while Figure~\ref{fig:surfacere_construction}(b) \& (c) show our results (see supplemental material for additional reconstruction results).
To reconstruct the surfaces, we follow the procedure in Sec.~\ref{sec:trainingandsynthesis} and employ the ball pivoting algorithm (BPA)~\cite{bernardini1999ball} to reconstruct the surfaces.
To show the effect of our network, we apply the same procedure with BPA to reconstruct surfaces directly from (i) the input point clouds (see Figure~\ref{fig:surfacere_construction}(d)), and from (ii) the input point clouds with PCA plane fitting for denoising (see Figure~\ref{fig:surfacere_construction}(e)), without  using our network to process the point clouds.

Comparing the resulting reconstructed surfaces with the ground truth meshes shown in Figure~\ref{fig:surfacere_construction}(f), we can see that our method achieves the best visual quality, particularly on preserving the edges.
In addition, we compute the mean Hausdorff distance between the reconstructed surfaces and their corresponding ground truth surfaces; see the mean errors shown in Figure~\ref{fig:surfacere_construction}(c), (d) \& (e); our method consistently has the lowest mean errors among all the testing models.
These quantitative results, together with the visual comparison, give evidence that our method produces consolidated point sets and improves the surface reconstruction quality.
Note that without the knowledge of edges learnt and recognized by the network, using PCA alone to denoise point clouds is not edge-aware; the sharp edges would be wiped away, if we directly apply PCA to the raw point cloud, leading to the inferior surface reconstructions shown in Figure~\ref{fig:surfacere_construction}(e).
To sum up, our consolidation facilitates high-quality reconstruction not only on sharp objects but also on usual smooth objects.
It is worth to note that with our consolidation, the overall reconstruction quality also improves over the state-of-the-art surface reconstruction methods on the benchmark models in~\cite{berger2013benchmark}; due to page limit, please see the supplemental material for more details.

{\bf Comparing with other methods.} \
\label{sec:comparewithothermethods}
In the experiment, we compare our method with state-of-the-art methods, EAR~\cite{huang2013edge}, CLOP~\cite{preiner2014continuous}, GPF~\cite{lu2018gpf}, and PU-Net~\cite{yu-2018-pu-net}, by applying them to process and consolidate point clouds before the screened Poisson surface reconstruction~\cite{kazhdan2013screened}.
As for PU-Net, we train a new model using our training objects and code released by the author.
For better comparison, we also apply patch-based manner in testing phase, as this can achieve better results.
Figures~\ref{fig:comparison1}(b), (c) \& (d) present the comparison with EAR and PU-Net.
We can see from the results that the sharp edges in the original mesh are mostly smoothed out if we take the point clouds from PU-Net for surface construction.
EAR better preserves the sharp features on the reconstructed surfaces, but in case of severe noise or under-sampling, it may over-smooth the geometric details and sharp features, or over-sharpen the edges.
It is because the method depends on the quality of the estimated normals; see the limitation paragraph in~\cite{huang2013edge}.
Our method, which is based on an edge-aware neural network model, is able to learn and capture edge information with high learning generalization, our point cloud consolidation results can lead to surface reconstructions that are closer to the original meshes.
Figure~\ref{fig:comparison2} also demonstrates that our method helps enhance the Poisson reconstruction quality on real scans in terms of preserving sharp edges compared with CLOP~\cite{preiner2014continuous} and GPF~\cite{lu2018gpf}.

%%%%%%%%%%%%%%%%%%%%%%%%%%%%%%%%%%%%%%%%%%%%%%
\newcommand{\BE}[1]{{\textbf{#1}}}
\begin{table*}[t]
	\caption{Quantitative comparison: our method, PU-Net, and EAR.}
	\label{tab:comparision1}
	\centering
	\begin{center}
		\begin{tabular}{c|c|c|c||c|c|c} \toprule[1pt]
			\multirow{2}*{Model}& \multicolumn{3}{c||}{Mean ($10^{-3}$)} & \multicolumn{3}{c}{RMS ($10^{-3}$)} \\
			\cline{2-7} 
			&Our		&PU-Net	&EAR		&Our	&PU-Net	&EAR  \\ \hline
			Camera	&\BE{1.31}	&1.91	&2.43		&\BE{1.99}	&2.74	&3.75 \\ \hline
			Sofa	&1.72		&3.20	&\BE{1.56}	&\BE{2.40}	&4.68	&2.87 \\ \hline
			Chair	&\BE{0.88}	&1.70	&1.93		&\BE{1.27}	&2.50	&3.54 \\ \hline
			Fandisk	&\BE{1.09}	&2.86	&2.33		&\BE{1.77}	&4.50	&5.63 \\ \hline
			Switch-pot	&\BE{1.36}	&2.00	&1.76		&\BE{1.83}	&3.07	&2.44 \\ \hline
			Headphone&\BE{0.81}&1.83	&3.71		&\BE{1.19}	&2.89	&6.93 \\ \hline
			Table	&\BE{1.15}	&2.14	&2.74		&\BE{1.62}	&3.12	&5.34 \\ \hline
			Monitor	&\BE{0.61}	&2.01	&2.58		&\BE{0.97}	&2.71	&5.73 \\ \bottomrule[1pt]	
		\end{tabular}
	\end{center}
\end{table*}

We also quantitatively compare with EAR and PU-Net by calculating the minimum distances from output points to the associated original mesh (as ground truth) in the test dataset.
Table~\ref{tab:comparision1} lists the mean and root mean square (RMS) values of different methods on the test models; see supplemental material for visual comparison.
We can see from the table that points generated from our method are closer to the original meshes compared to others.
The PU-Net uses the EMD loss to encourage output points to be located on the underlying surfaces, so this comparison shows that the results are sub-optimal compared with our method, which directly minimizes the distances from output points to surfaces.

\begin{figure}[!t]
	\centering
	\includegraphics[width=0.85\linewidth]{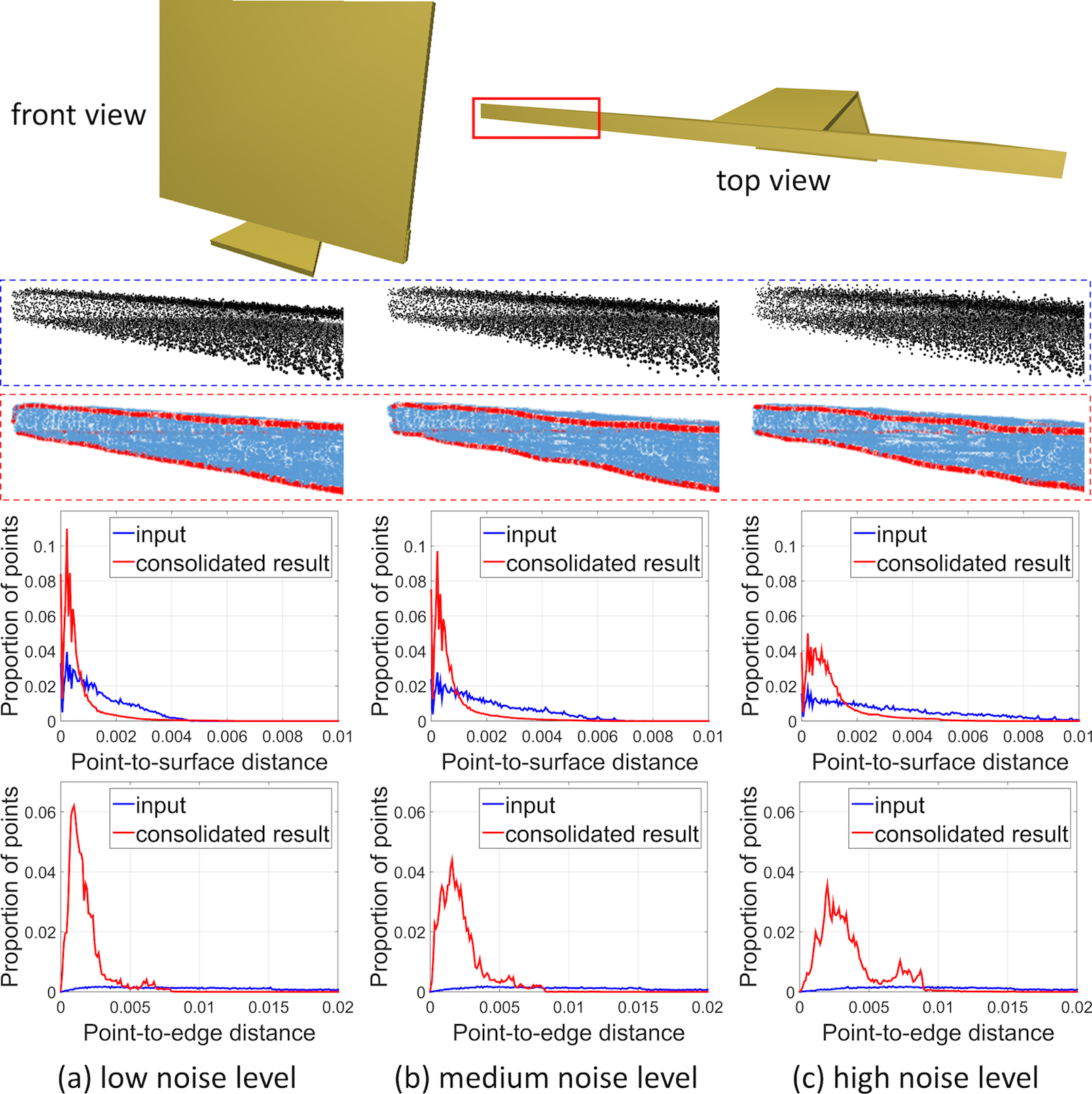}
	\caption{Results of using our method to consolidate point clouds with different amount of quantization noise.
		Top row: the testing Monitor model;
		2nd and 3rd rows: blown-up views of the input point clouds with different noise level (low to high) and our consolidated results, respectively;
		4th and 5th rows: statistics of point-to-surface and point-to-edge distances, respectively.}
	\label{fig:noise_robustness}
\end{figure}

\begin{figure*}[!t]
	\centering
	\includegraphics[width=0.9\linewidth]{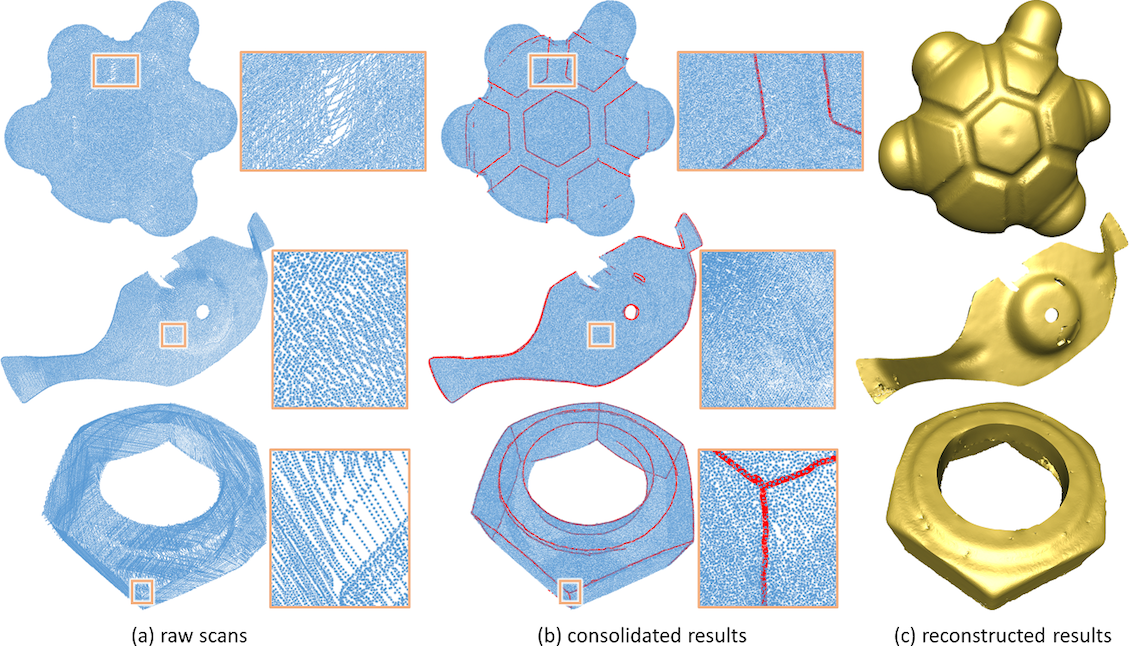}
	\caption{Three results from real scans (a).
		Note the diversity of the geometry, the sparseness of the inputs (a), and how well the network locates the edges (in red); see (b).
		The reconstruction results (c) are yet imperfect due to tiny regions that are severely undersampled (see the blown-up views in (a)).}
	\label{fig:real_scan}
	\vspace*{-10pt}
\end{figure*}

{\bf  Our results under varying quantization noise.} \
\label{sec:noise}
In real scans, the acquired depth values are generally quantized nonlinearly depending on the distance from the scanner.
In short, objects far from the scanner would have less precise depth values in fewer quantization levels.
During the virtual scanning process, we mimic real scans by quantizing the depth values in $N_q$ levels.
This means that there are only $N_q$ unique depth values over all the foreground pixels.
In this subsection, we investigate the robustness of our method under different amount of quantization noise by producing three sets of point cloud data using virtual scan on a testing model with $N_q$$=$$120$ (low noise), $N_q$$=$$80$ (medium noise), and $N_q$$=$$50$ (high noise) and then applying our method to consolidate the resulting point clouds.

Figure~\ref{fig:noise_robustness} presents the results, where the left, middle and right columns correspond to low, medium and high noise levels, respectively.
From the blown-up views, we can see more points above the object surface for the point cloud with high noise.
The statistics about the point-to-surface distances also reveal the noise level; see the blue plots in the 4th row.
After the consolidation, the points are steered to the left (see the red plots in 4th row), meaning that the points are now all closer to the ground truth surface under different noise levels.
Similar pattern can also be observed from the statistical results for the point-to-edge distances.

{\bf Results on real scans.} \
\label{sec:real_scans}
We also apply our method to point clouds produced from real scans downloaded from Aim@Shape and obtained from the EAR project~\cite{huang2013edge}.
Figure~\ref{fig:comparison2} has shown some results on real scans, and Figure~\ref{fig:real_scan} shows more consolidation and reconstruction results.
As we can see, real scan point clouds are often noisy and have inhomogeneous point distribution.
Comparing with the input point clouds, our method is still able to generate more points near the edges and on the surface, while better preserving the sharp features.

%%%%%%%%%%%%%%%%%%%%%%%%%%%%%%%%%%%%%%%%%%%%%%%%%%%%%%%%%%%
\section{Discussion and future works}
\label{sec:conclusion}

We presented EC-Net, the first edge-aware network for consolidating point clouds.
The network was trained on synthetic data and tested on both virtual and real data.
To be edge-aware, we design a joint loss function to identify points along edges, and to encourage points to be located close to the underlying surface and edges, as well as being more evenly distributed over surface.
We compared our method with various state-of-the-art methods for point cloud consolidation, showing improvement in the 3D reconstruction quality, especially at sharp features.
Our method has a number of limitations.
First, it is not designed for completion, so filling large holes and missing parts would be a separate and different problem.
In future, we plan to investigate how to enhance the network and the training process for point cloud completion.
For tiny structures that are severely undersampled, our network may not be able to reconstruct the sharp edges around them.
With insufficient points, the patch could become too large compared to the tiny structure.
Moreover, our current implementation considers patches with a fixed number of points, and it cannot adapt structures of varying scales.
It would be an interesting problem to try to dynamically control the patch size and explore the use of larger context in the training and inference.

{\bf Acknowledgments.} \
We thank anonymous reviewers for the comments and suggestions. The work is supported by the Research Grants Council of the Hong Kong Special Administrative Region (Project no. GRF 14225616), the Shenzhen Science and Technology Program (No. JCYJ20170413162617606 and No. JCYJ20160429190300857), and the CUHK strategic recruitment fund.

\clearpage
\bibliographystyle{splncs}
\bibliography{sharppts}

%%%%%%%%%%%%%%%%%%%%%%%%%%%%%%%%%%%%%%%%%%%%%%%%%%%%%%%%%%%%%%%%%%%%%%%%%%%%%%
\newpage
\appendix

\section{Our Training Data set}
\label{supp_dataset}

\begin{figure}[!h]
	\centering
	\vspace*{5mm}
	\includegraphics[width=1\linewidth]{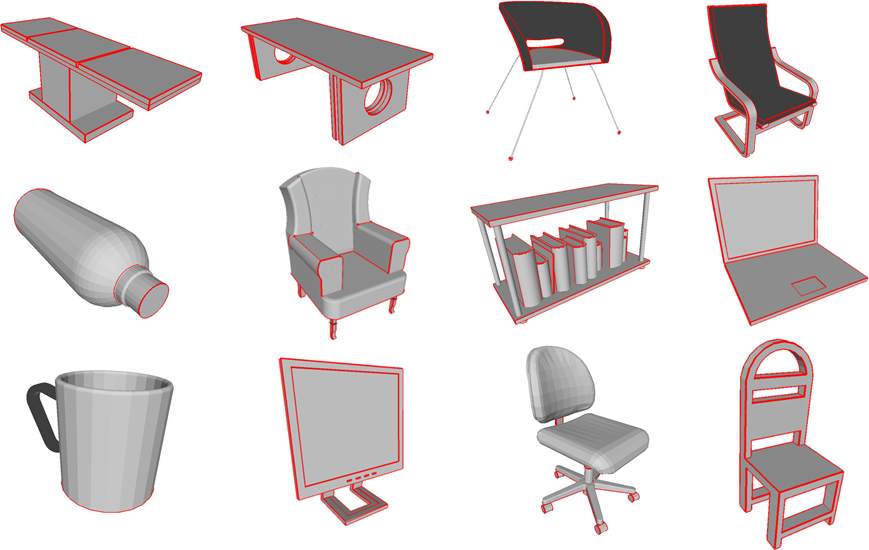}
	\caption{Training models (12 everyday objects) with the annotated edges marked in red.}
	\label{fig:dataset_train2}
\end{figure}

\begin{figure}[!h]
	\centering
	\includegraphics[width=0.95\linewidth]{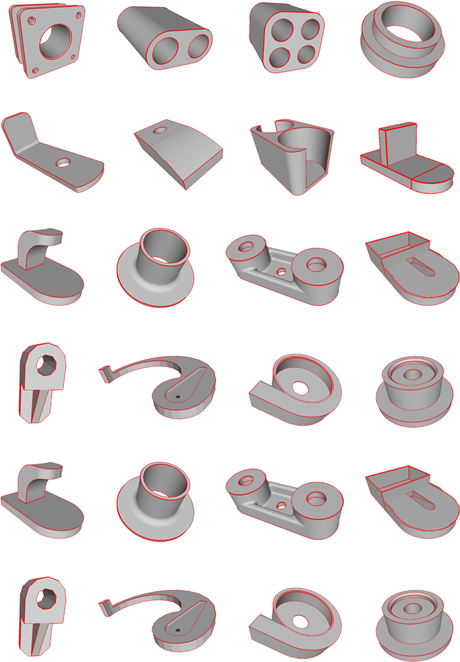}
	\caption{Training models (24 CAD models) with the annotated edges marked in red.}
	\label{fig:dataset_train1}
\end{figure}

\noindent
We collected 12 everyday objects and 24 CAD models in total as our training data set, and manually annotate sharp edges on the models.
Figures~\ref{fig:dataset_train2} and~\ref{fig:dataset_train1} show the training models with the annotated edges. 
%We also present the shapes of some training and testing 3D models in our dataset in Fig.~\ref{fig:dataset_train1} and Fig.~\ref{fig:dataset_train2}, respectively.  
%As we can see, our collected datasets have a large variation in geometry shapes, containing 3D models with smooth surface regions (first row) and 3D models with sharp corners and edges (second row). 
%There is also a large variation between training and testing 3D models, indicating a good generalization ability of our proposed method. 

\clearpage

%%%%%%%%%%%%%%%%%%%%%%%%%%%%%%%%%%%%%%%%%%%%%%%%%%%%%%%%%%%%%%%%%%%%%%%%%%%%%%

\section{Network Architecture of EC-Net}
\label{supp_network}

\begin{figure}[!h]
	\centering
	\includegraphics[width=1.0\linewidth]{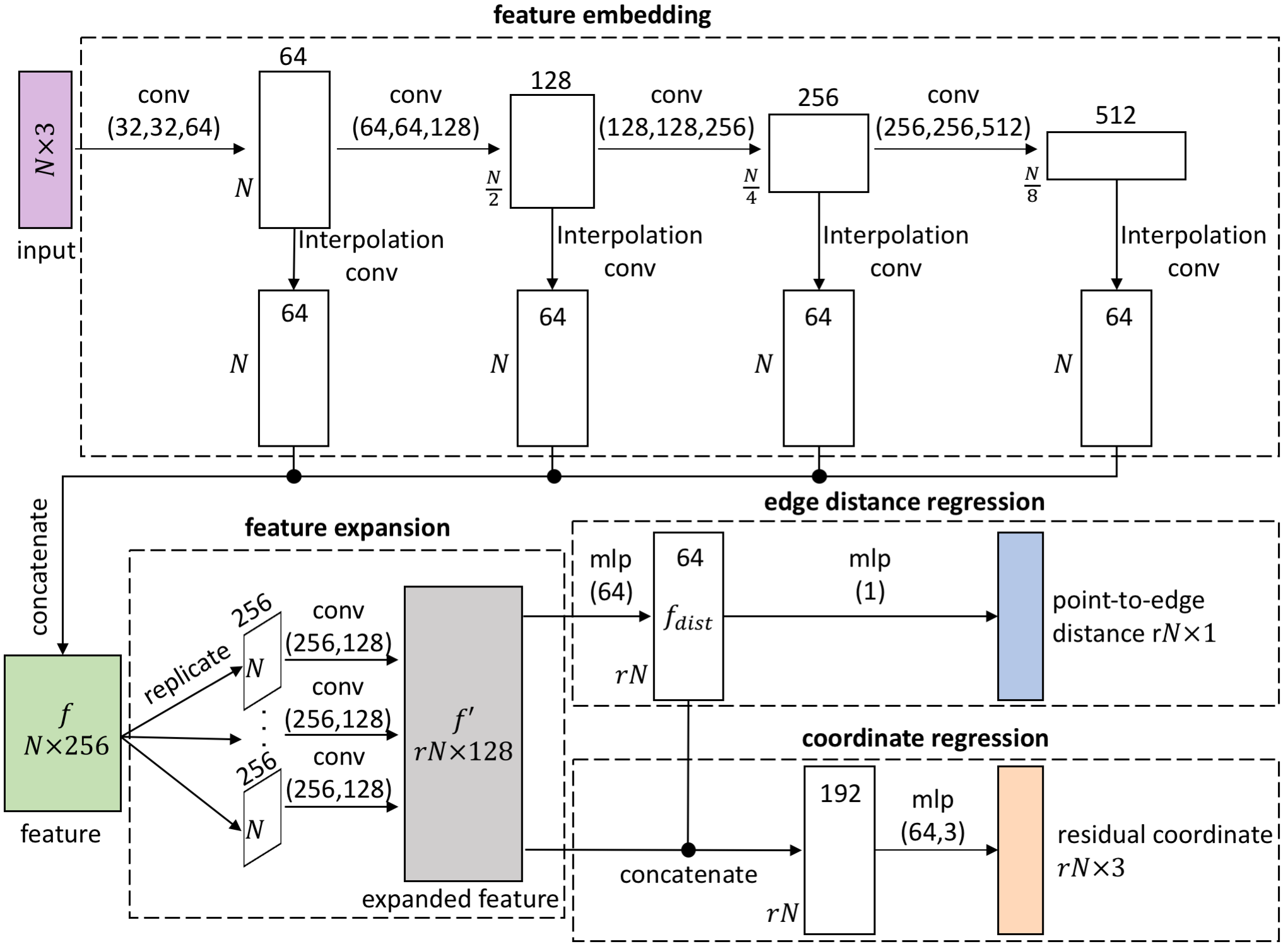}
	\vspace*{-1.5mm}
	\caption{The network architecture of EC-Net.}
	\label{fig:arch}
	\vspace*{-2mm}
\end{figure}

\noindent
Figure~\ref{fig:arch} presents the network architecture of EC-Net; see the explanation below for the details:
%The details of our network architecture are listed as follows.
%
\begin{itemize}
	\item
	The feature embedding component (see the dashed box on top) is based on PointNet++, where we adopt four levels with grouping radii 0.1, 0.2, 0.4, and 0.6 to extract the local features.
	The corresponding number of point samples in these four levels are $N$, $\frac{N}{2}$, $\frac{N}{4}$, and $\frac{N}{8}$, respectively.
	\item
	Next, we directly concatenate features from different levels to aggregate the multi-level feature.
	Specifically, for each level, we use the interpolation in PointNet++ to restore the feature of the level, and then use a convolution to reduce the restored feature to 64 dimensions.
	After that, we concatenate the restored features from the four levels to form the $D$ = 256 dimensional feature denoted as $f$ (see the green box above).
	%
	%\phil{For each word ``interpolation'' in the figure above, shall we say ``interpolation \& convolution'' instead of just ``interpolation''?}
	%
	\item 		
	In the feature expansion component (see the dashed box on lower left), we follow the feature expansion module in PU-Net.
	Specifically, we create $r$ copies of feature $f$, independently apply two separated convolutions with 256 and 128 output channels to each copy, and then concatenate the outputs to form the expand feature denoted as $f'$ (see the grey box above).
	\item
	In the edge distance regression component, we first use one fully-connected layer with width 64 to regress $f_{dist}$ from $f'$.
	Then, we use another fully-connected layer with width 1 to regress the point-to-edge distance $d$.
	\item 
	In the coordinate regression component, we first concatenate the $f_{dist}$ and $f'$ to form a 192 dimensional feature.
	Then, we use two fully-connected layers with width 64 and 3 to regress the residual point coordinates from the concatenated feature, and output the final 3D coordinates by adding back the original point coordinates.
\end{itemize}
All the convolution layers and fully-connected layers in the above network are followed by the ReLU operator, except for the last two point-to-edge distance and residual coordinate regression layers.

\if 0
\begin{figure}[t]
	\centering
	\includegraphics[width=0.88\linewidth]{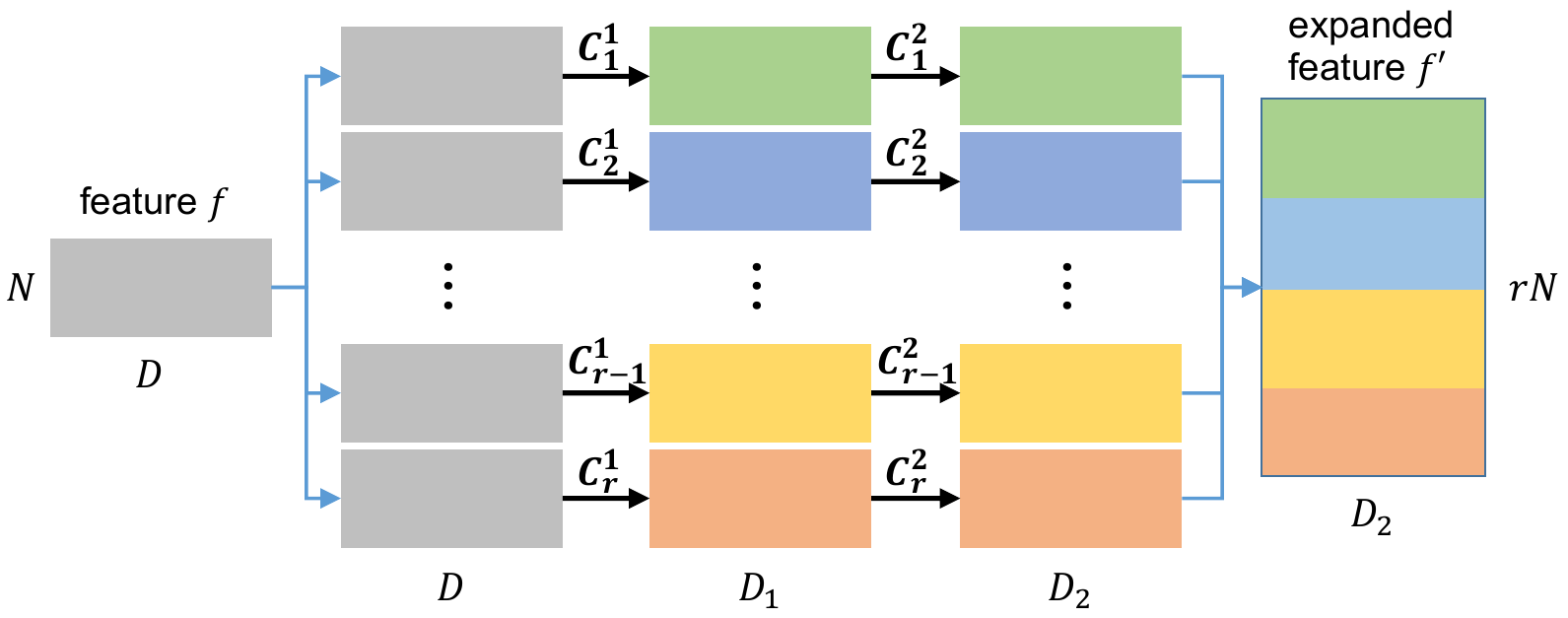}
	\vspace*{-1.5mm}
	\caption{The feature expansion component.
		The input is feature $f$ (dimension: $N$$\times$$D$) from the feature embedding component, and the output is expanded feature $f'$ (dimension: $rN$$\times$$D_2$), where
		$N$ is the number of points,
		$r$ is the upsampling rate, and
		$D$, $D_1$ and $D_2$ are feature dimensions.
		%%
		First, we replicate $f$ to $r$ copies, and apply two separate convolutions $C^1_i$ and $C^2_i$ to each copy.
		Then, we concatenate the $r$ sets of convolution features along the first dimension as the output.}
	\label{fig:featureexpansion}
	\vspace*{-2mm}
\end{figure}
\fi 

%%%%%%%%%%%%%%%%%%%%%%%%%%%%%%%%%%%%%%%%%%%%%%%%%%%%%%%%%%%%%%%%%%%%%%%%%%%%%%
\clearpage
\section{More Experimental Results}
\label{supp_result}

%%%%%%%%%%%%%%%%%%%%%%%%%%%%%%%%%%%%%%%%%%%%%%%
\subsection{Comparison with other reconstruction methods on benchmark models}

\noindent
Besides the comparison in the main paper, we further evaluate our method, and compare with other surface reconstruction methods on the reconstruction benchmark models~\cite{berger2013benchmark}.
We also include a recent method for large-scale surface reconstruction~\cite{UB17}. 

Fig.~\ref{fig:benchmark} below shows the comparison results on two sharp objects in the benchmark, where (b)-(d) are results downloaded from the benchmark's website related to the best three visual methods presented in the website.
%%
%%Specifically, we directly use our well-trained model to consolidate the more noisy point cloud and then do reconstruction. 
%%
The comparison clearly shows that the reconstructions with our consolidation better preserve the sharp edges and have better visual quality even on the noisy models from the benchmark dataset with random error and systematic error.

\begin{figure}[h]
	\centering
	\includegraphics[width=1.0\linewidth]{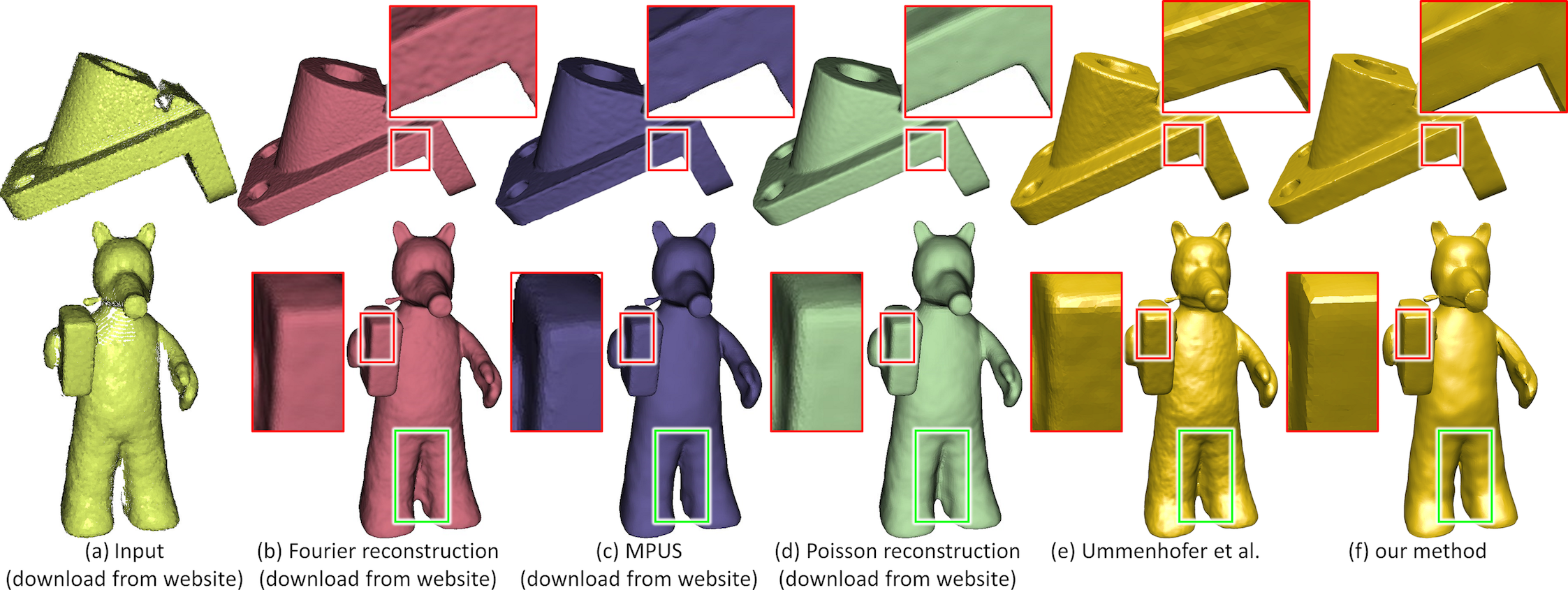}
	\caption{Comparison of different reconstruction methods on benchmark models.}
	\label{fig:benchmark}
\end{figure}

%%%%%%%%%%%%%%%%%%%%%%%%%%%%%%%%%%%%%%%%%%%%%%%%%%%
\clearpage
\subsection{Additional point consolidation and surface reconstruction results}

\begin{figure}[h]
	\centering
	\includegraphics[width=1.0\linewidth]{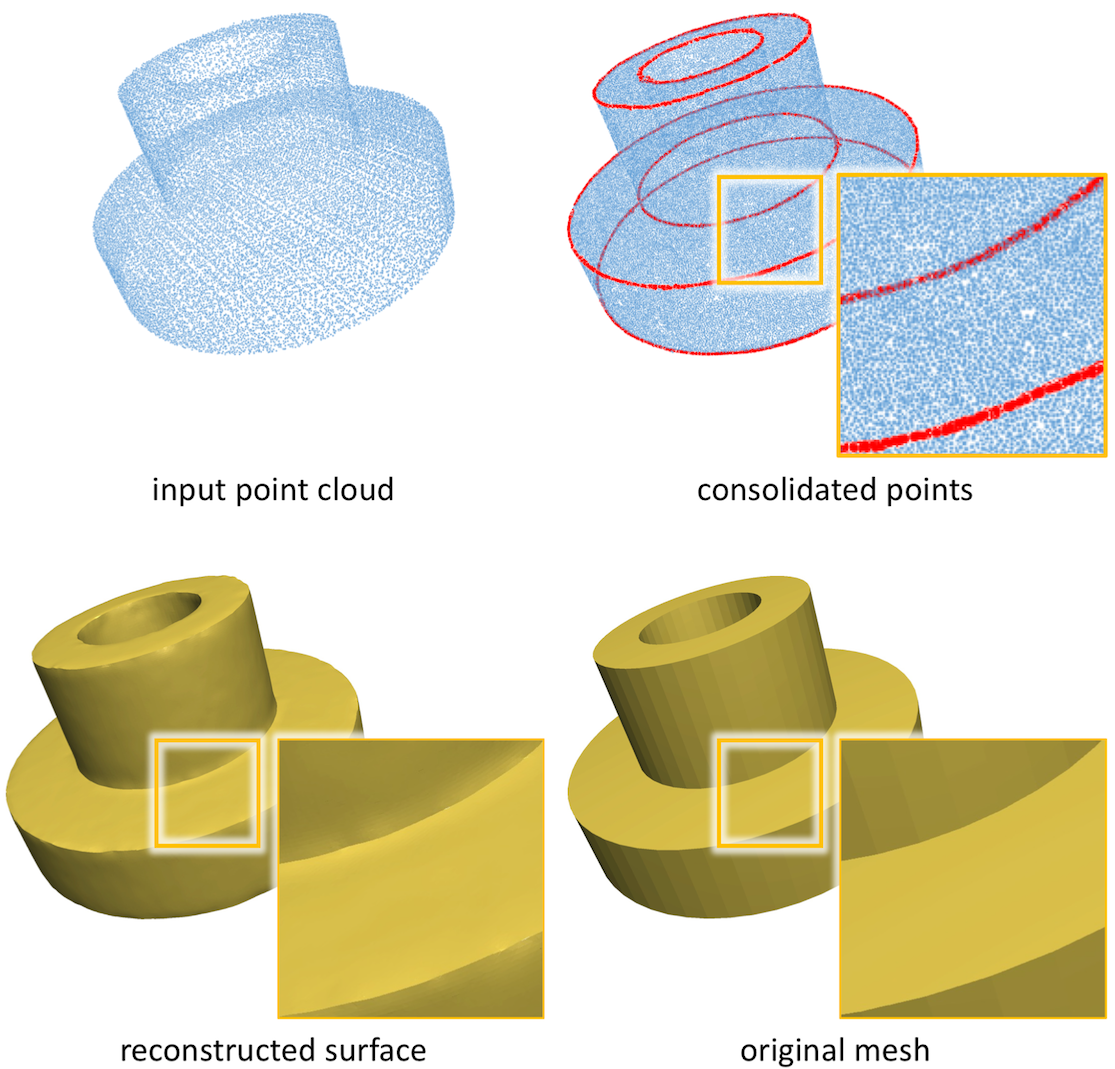}
	\caption{Point consolidation and surface reconstruction results produced with our method.
		%\phil{try to better align the subimages: lower right subimage is a bit higher than lower left subimage}
	}
	\label{fig:surfacereconstruction1}
\end{figure}

\begin{figure}[h]
	\centering
	\includegraphics[width=1.0\linewidth]{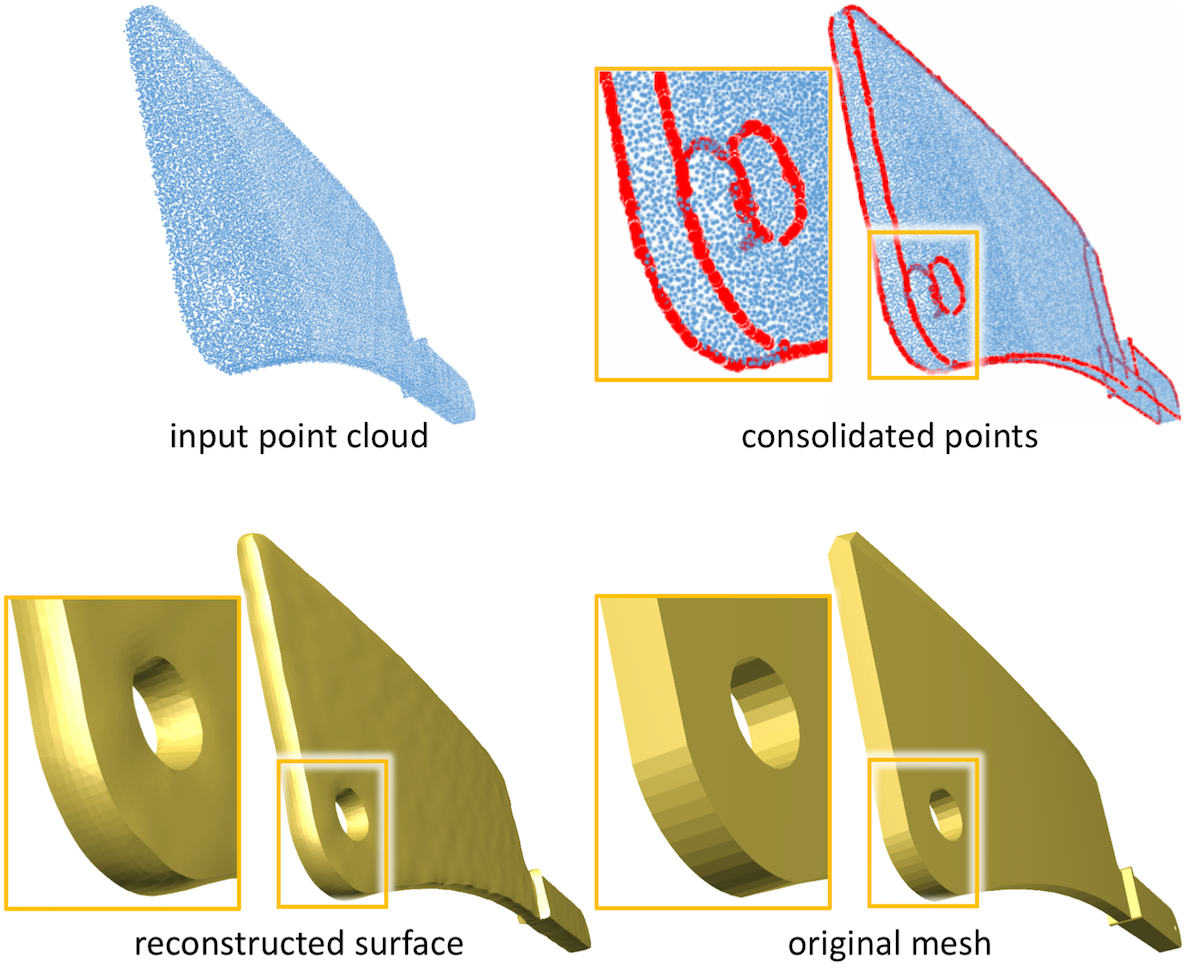}
	\caption{Point consolidation and surface reconstruction results produced with our method.
		%\phil{make each subimage bigger... until margin?}
	}
	\label{fig:surfacereconstruction2}
\end{figure}

\begin{figure}[h]
	\centering
	\includegraphics[width=1.0\linewidth]{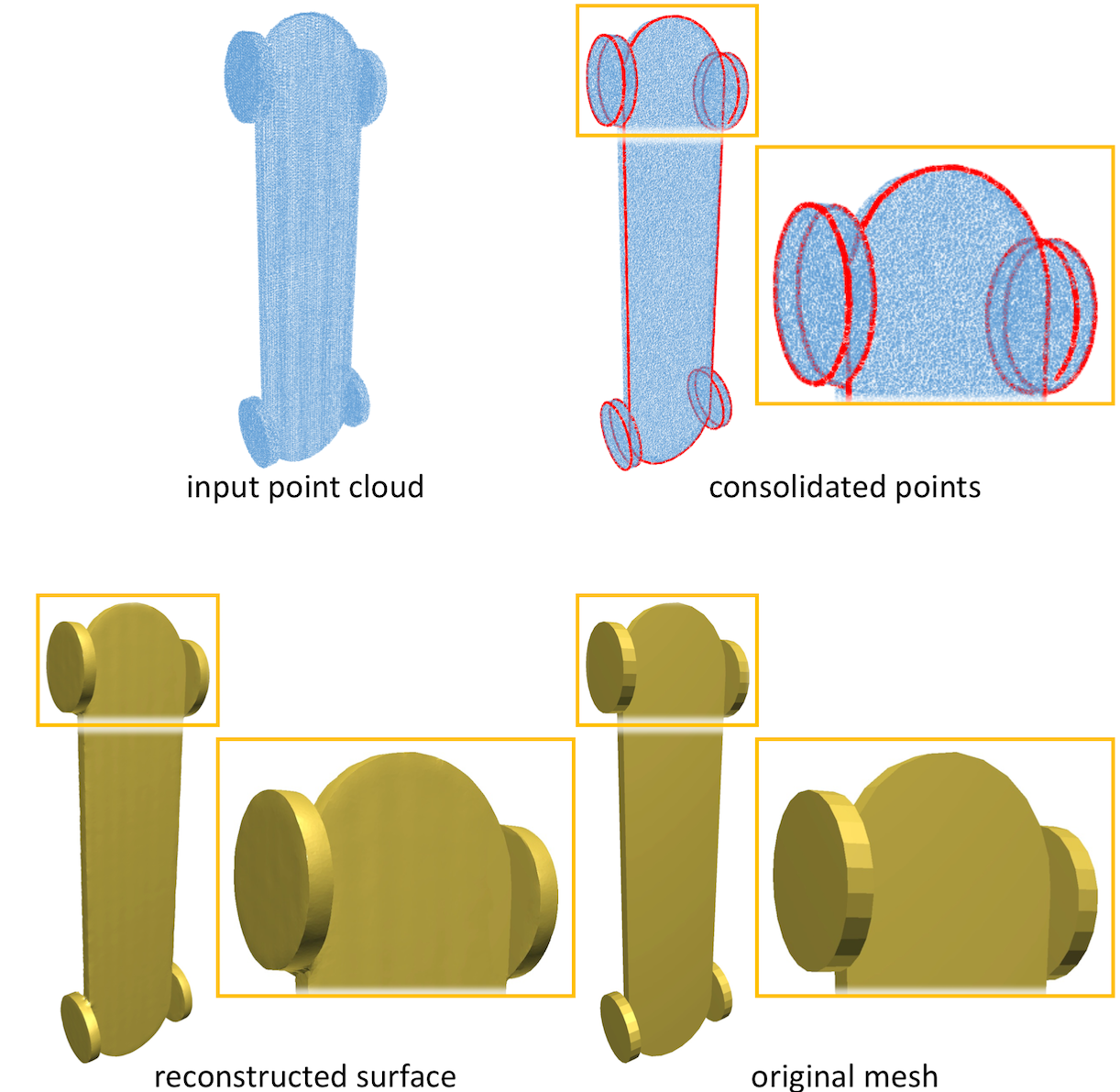}
	\caption{Point consolidation and surface reconstruction results produced with our method.}
	\label{fig:surfacereconstruction3}
\end{figure}

\noindent
Figures~\ref{fig:surfacereconstruction1} to~\ref{fig:surfacereconstruction3} present additional point consolidation and surface reconstruction results produced with our method.

%%%%%%%%%%%%%%%%%%%%%%%%%%%%%%%%%%%%%%%%%%%%%%%%%%%
\clearpage
\subsection{Visual comparison of point-to-surface distances}

\begin{figure}[h]
	\centering
	\includegraphics[width=1.0\linewidth]{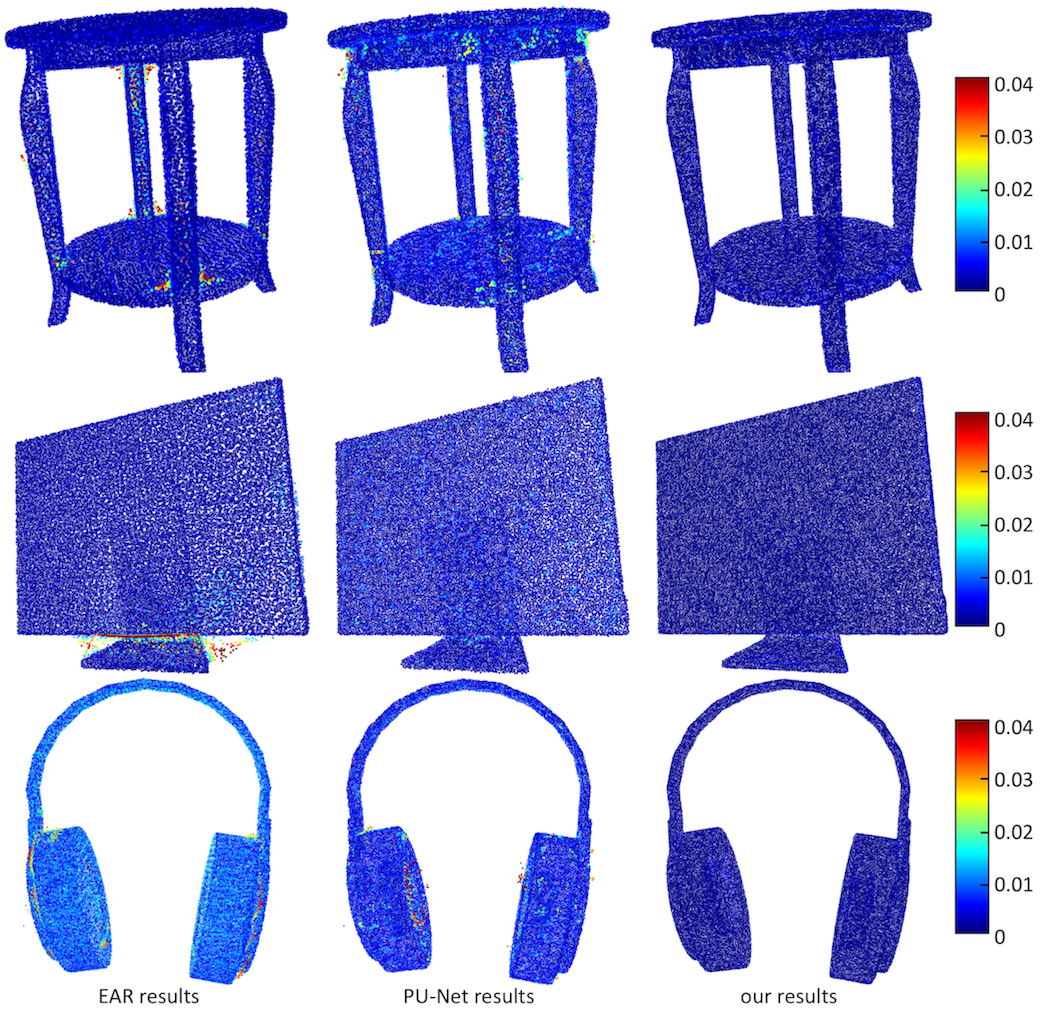}
	\caption{Visual comparison of point-to-surface distances for results produced with EAR, PU-Net and EC-Net.
		Each point is colored according to its minimum distance from the original (ground truth) mesh surface.}
	\label{fig:visualerror}
\end{figure}

\noindent
Figure~\ref{fig:visualerror} presents visual comparison of point-to-surface distances for results produced with EAR, PU-Net and our EC-Net on three different models (see also Table 1 in the submitted paper).
In detail, we color each point based on its minimum distance (proximity) to the original mesh, i.e., the ground truth surface; see the color maps on the right.
From these results, we can see that points in results produced with EC-Net are clearly darker, i.e., having lower point-to-surface distances.

%%%%%%%%%%%%%%%%%%%%%%%%%%%%%%%%%%%%%%%%%%%%%%%%%%%
\clearpage
\section{Surface Reconstruction Results from Point Clouds of Varying Number of Points}
\label{sec:sparse]}

%we conduct the following experiment as shown in Figure~\ref{fig:sparsetest}

\begin{figure}[h]
	\centering
	\includegraphics[width=1.0\linewidth]{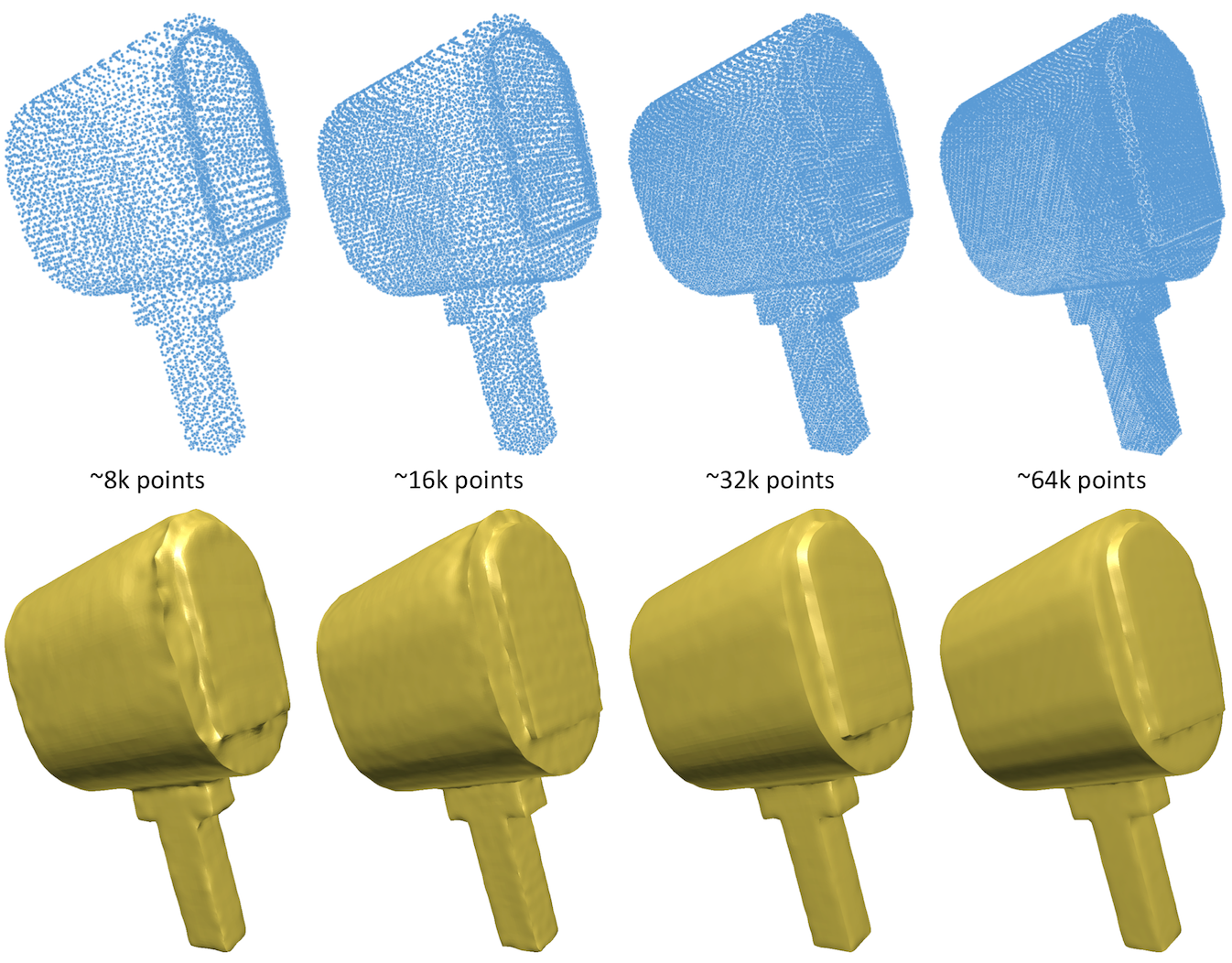}
	\caption{Surface reconstruction results (bottom) produced from point clouds of different number of points (top).}
	%; from left to right: ~8k, ~16k, ~32k and ~64k).}
	\label{fig:sparsetest}
\end{figure}

\noindent
To study the ability of our method to handle input point clouds of varying number of point samples, we virtually scan the same 3D model to produce four point clouds of around 8k, 16k, 32k, and 64k point samples; see the top row in Figure~\ref{fig:sparsetest}.
Then, we apply our method to each point cloud and produce a surface reconstruction result accordingly; see the bottom row in the figure.
From the results, we can observe that for the input point cloud with only $\sim$8k points, the point distribution is severely non-uniform and inhomogeneous, with some of the points being very close to one another.
For such a severe case, our method can still help reconstruct the surface and edges, yet the surface may not be very smooth.
%With increasing number of input points, the quality of the reconstructed surface improves. 

\end{document}